\documentclass[final]{cvpr}

\usepackage{times}
\usepackage{epsfig}
\usepackage{graphicx}
\usepackage{amsmath}
\usepackage{amssymb}

\usepackage{siunitx}
\usepackage{wrapfig}
\usepackage{multirow}
\usepackage{booktabs} % To thicken table lines
\usepackage{float}
\usepackage[numbers,sort]{natbib}
\usepackage{pifont}% http://ctan.org/pkg/pifont
\usepackage{cuted}
\usepackage{capt-of}

\newcommand{\cmark}{\ding{51}}%
\usepackage[pagebackref=true,breaklinks=true,colorlinks,bookmarks=false]{hyperref}

\def\cvprPaperID{5540} % *** Enter the CVPR Paper ID here
\def\confYear{CVPR 2021}

\begin{document}

%%%%%%%%% TITLE
\title{Shelf-Supervised Mesh Prediction in the Wild}
\vspace{-5mm}
\author{Yufei Ye\textsuperscript{1} \qquad Shubham Tulsiani\textsuperscript{2} \qquad Abhinav Gupta\textsuperscript{12}  \\
\textsuperscript{1}Carnegie Mellon University  \qquad \textsuperscript{2}Facebook AI Research \\
{\tt \small yufeiy2@cs.cmu.edu \qquad \{shubhtuls,gabhinav\}@fb.com}
\\
{\tt \small \href{https://judyye.github.io/ShSMesh/}{https://judyye.github.io/ShSMesh/}}
}

\maketitle

% \begin{figure*}[h!]
%     \centering
%     \includegraphics[width=\linewidth]{exp/teaser.png}
%   \caption{Given a single image, we predict a mesh with textures, rendered from a novel view and the predicted view. \todo{judy: 2 col figure, 1 / each classes? }}
%      \label{fig:teaser}
% \end{figure*}

\begin{strip}\centering
\vspace{-1.7cm}
\includegraphics[width=\textwidth]{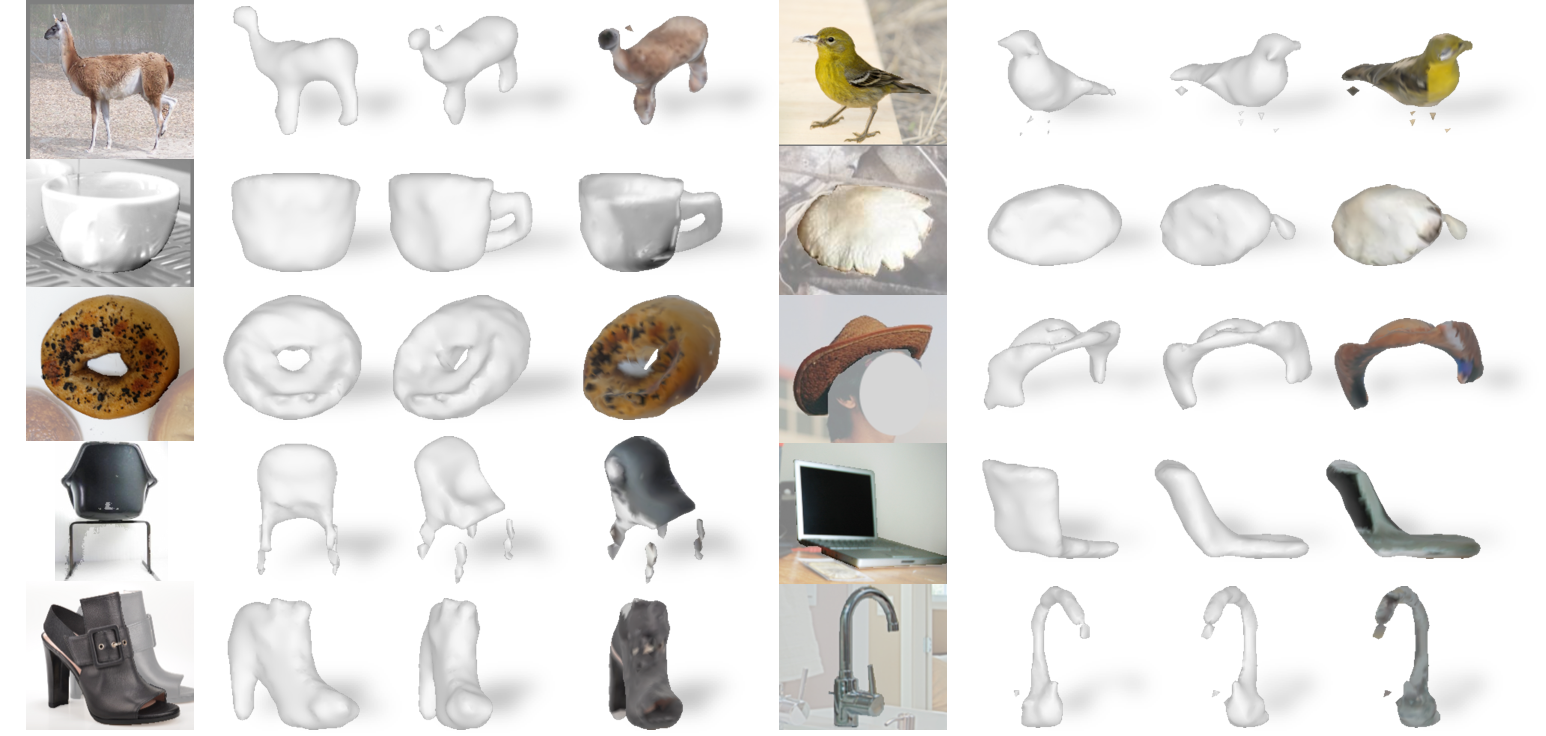}
\captionof{figure}{Given a single image, we predict a mesh with textures (rendered from the predicted view and a novel view). The models can learn directly from collections of images with only foreground masks, without supervision of mesh templates, multi-view association, camera poses, semantic annotations, \etc.
\label{fig:teaser}}
\vspace{-.2cm}
\end{strip}

% \begin{wrapfigure}{r}{0.35\textwidth}
%   \begin{center}
% \vspace{-4.5em}
%     \includegraphics[width=0.35\textwidth]{fig/teaser.png}
%   \end{center}
%   \vspace{-1.5em}
%   \caption{Given a single image, we predict a mesh with textures, rendered from a novel view and the predicted view.}
% %   \vspace{-2em}
% \end{wrapfigure}

\begin{abstract}
\vspace{-5mm}
We aim to infer 3D shape and pose of object from a single image and propose a learning-based approach that can train from unstructured image collections, supervised by only segmentation outputs from off-the-shelf recognition systems (\ie `shelf-supervised'). We first infer a volumetric representation in a canonical frame, along with the camera pose. We enforce the representation geometrically consistent with both appearance and masks, and also that the synthesized novel views are indistinguishable from image collections. The coarse volumetric prediction is then converted to a mesh-based representation, which is further refined in the  predicted camera frame. These two steps allow both shape-pose factorization from image collections and per-instance reconstruction in finer details. We examine the method on both synthetic and the real-world datasets and demonstrate its scalability on 50 categories in the wild, an order of magnitude more classes than existing works.
 
% We report performance on both synthetic and real world datasets. Experiments show that our approach captures category-level 3D shape from image collections more accurately than alternatives, and that this can be further refined by our instance-level specialization. 
\end{abstract}

\section{Introduction}
\vspace{-2mm}
We live in a 3D world where 3D understanding plays a crucial role in our visual perception. Yet most computer vision systems in the wild still perform 2D semantic recognition (classification/detection). Why is that? We believe the key reason is the lack of 3D supervision in the wild. Most recent advances in 2D recognition have come from supervised learning but unlike 2D semantic tasks, obtaining supervision for 3D understanding is still not scalable.

While some recent approaches~\cite{gkioxari2019mesh, wu2016learning} have attempted to build supervised 3D counterpart of 2D approaches, the concerns about scalability still remain. Instead, a more promising direction is to learn models of single image 3D reconstruction by minimizing the amount of manual supervision needed. Early approaches in this direction focused on using multi-view supervision~\cite{wiles2017silnet, yan2016perspective}. However, obtaining multiple views of the same objects/scene is still not easy for the data in the wild. Therefore, recent approaches~\cite{cmrKanazawa18, kulkarni2019csm,nguyen2019hologan} have attempted to learn single-image 3D reconstruction models from image collections. These approaches have targeted use of category templates, pose supervision and keypoints to provide supervision (See Table~\ref{tab:sup}). However, such supervision still limits the scalability to hundreds of categories. 

Our work is inspired by recent approaches that forgo supervision by exploiting meta-supervision from the category structure and geometric nature of the task. More specifically, the two common supervisions used are: (a) {\bf rendering supervision} (\cite{kulkarni2019csm, li2020self}): any given image of an instance in a category is merely a rendering of a 3D structure under a particular viewpoint. We can therefore enforce that the inferred 3D shape be consistent with the available image evidence when rendered; (b) {\bf adversarial supervision} (\cite{nguyen2019hologan}): in addition, the availability of an image collection also allows us to understand what renderings of 3D structures should look like in general. This enables us to derive supervisory signal not just from renderings of predictions in the input view, but also from novel views, by encouraging the novel-view renderings to look realistic. Prior work has exploited these supervisions but individually they pose several limitations for scaling 3D reconstruction models. For example, \cite{kulkarni2019csm, goel2020shape} still requires template models. Similarly, \cite{nguyen2019hologan} exploits the adversarial supervision and ignores the explicit geometric supervision. Therefore, such an approach only works on categories with strong structure and curated image collections.

This paper attempts to build upon the very recent successes in meta-supervision and provide an approach to scale learning of single image 3D reconstruction in the wild. We present a two-step approach: the first step relies on category-level understanding for coarse 3D inference (learned via meta-supervision). The second step specializes coarse models to match the details in the input image. Our approach can learn using only unannotated image collections, without requiring any ground-truth 3D \cite{bogo2016keep, wang2018pixel2mesh, gkioxari2019mesh}, multi-view \cite{drcTulsiani19, yan2016perspective}, category templates \cite{kulkarni2020acsm, goel2020shape}, or pose  supervision \cite{wiles2017silnet, kato2019learning}. This not only allows our approach to infer accurate 3D, but also enables it to do so beyond the  synthetic settings, using in-the-wild image collections in a `shelf-supervised' manner: with only approximate instance segmentation masks obtained using off-the-shelf recognition systems as supervision. Yet our biggest contribution is the demonstration of scalability -- we show results on order of magnitude more classes than existing papers.

\section{Related Work}

\noindent\textbf{Supervised Single-view 3D Reconstruction. }  
% There is  inherent shape-pose ambiguity  to infer 3D structure and pose from a single image. To reduce the ambiguity, the main idea is to introduce priors  
% To resolve the inherent ambiguity in this task, 
When considering object categories that exhibit limited shape variation, a common approach is to leverage predefined 3D deformable templates to either fit the input \cite{bogo2016keep, xiang2019monocular, habermann2019livecap, pepik2012teaching} or to regress the model parameters \cite{wang2018pixel2mesh, su2015render, smith2019geometrics}. While they achieve remarkable performance to reconstruct details, the predefined shape model may not be available for an arbitrary category or one template is not sufficient to fit all instances with varying topologies and shapes (\eg chairs). Another line of works \cite{akhter2015pose, wu2016learning, girdhar2016learning, park2019deepsdf, sitzmann2019scene} learns a manifold of shape by first mapping the input to a latent space from which 3D shape is generated. However, these methods suffer from losing finer details as the reconstruction only rely on less expressive latent code. 
% as previous work \cite{tatarchenko2019single} argues, . 
Similar to our approach, Gkioxari \etal \cite{gkioxari2019mesh}  recently addressed these concerns by first inferring a coarse shape, and then refining it to match the given image. While all methods above have presented impressive results, they crucially require 3D supervision. In contrast, our work aims for similar inference where neither 3D  nor pose annotation is available.

% Our work learns to initialize the template directly from the observed data rather than querying external knowledge. 

% Apart from those, manual priors are also used such as symmetry \cite{}, laplacian smoothness \cite{}, low-dimension \cite{}, 

\begin{table}[]
    \caption{Comparing ours  to other image-based supervised works  in terms of supervision and outputs.}
    \centering
    \footnotesize
    \begin{tabular}{r |  p{.5em} p{.5em} p{.5em} p{.5em} p{.5em} p{.5em} p{.5em} p{.5em} p{.5em}   c }
% &\cite{gkioxari2019mesh} \cite{choy20163d} 
& \cite{henderson2020leveraging} & \cite{kulkarni2020acsm} & \cite{cmrKanazawa18}  & \cite{kato2019learning} & \cite{drcTulsiani19}  & \cite{Wu_2020_CVPR}&  \cite{nguyen2019hologan} & \cite{gadelha2017prgan} & \cite{li2020self}  &  ours \\
\hline
% 3D-free & \xmark &  & & & & & & &\\
pose  & \cmark & & \cmark & \cmark & &  & & &\\
template  & & \cmark & \cmark & \cmark & & & & & \\
semantic  & & & \cmark &  & & & & & (\cmark) &  \\
multi-view  & & &  & & \cmark &  & & & & \\
mask & (\cmark) & \cmark & \cmark & \cmark & \cmark &   & & \cmark & \cmark & \cmark\\
\hline
3D recon. & & \cmark & \cmark  & \cmark & \cmark & (\cmark) & & \cmark & \cmark  & \cmark \\
topology  & &  & & & \cmark & &  & \cmark &  & \cmark \\
texture & \cmark & & \cmark & \cmark & \cmark & \cmark & \cmark & & \cmark & \cmark
    \end{tabular}
    \label{tab:sup}
\vspace{-5mm}
\end{table}
\noindent\textbf{Unsupervised learning from image-based supervision.} 
With a similar motivation as ours to relax the need of supervision, several approaches study the reconstruction task with only multi-view or even single-view supervision. 
The key is to ensure reprojection consistency of the predicted 3D with available observations. While this relaxes the requirement for tedious 3D supervision, manual annotations are still required in different forms, such as  semantic key-points \cite{cmrKanazawa18, kong2019deep}, multi-view association \cite{drcTulsiani19, choy20163d, wiles2017silnet, liu2019learning}, categorical template \cite{cmrKanazawa18, kulkarni2019csm, kulkarni2020acsm, goel2020shape}, or camera pose annotation \cite{kato2019learning, yan2016perspective, henderson2019learning, henderson2020leveraging, liu2019learning}. 
Some recent works use self-supervised semantic co-part segmentation \cite{li2020self}, foreground masks \cite{gadelha2017prgan, henzler2019escaping}, or symmetry \cite{Wu_2020_CVPR} to further relax the manual annotation. Our work has similar setup while ours  does not require semantic in training, and reconstructs textured full 3D meshes with various topology and shapes.  Table \ref{tab:sup} summarizes the differences of our method with others in terms of supervision and outputs.

% Some recent work \cite{li2020self} uses automatically generated semantic segmentation but the reconstruction is limited to genus-0 surfaces.   In contrast, our work only requires foreground masks, which can even be automatically obtained, and is able to reconstruct shapes with various topology. Perhaps the most relevant to our learning setup are  \cite{gadelha2017prgan, henzler2019escaping} that demonstrate encouraging results. 
% The formal use only silhouette masks and the later leverages an explicit volume-based renderers.  
% While these approaches predict plausible volumetric representation, we aim to predict meshes that are specialized to the input image. 

% \todo{text: shangzhe, contrast to platogan, li .}

\begin{figure*}
    \centering
    \includegraphics[width=\textwidth]{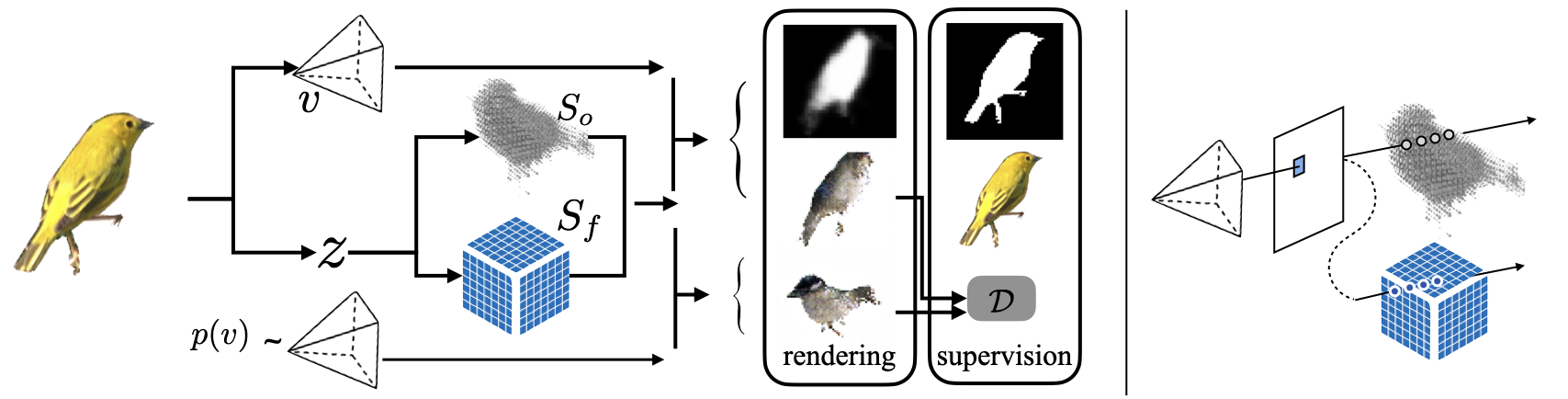}
    \caption{\textbf{Volumetric Representation Prediction and Rendering.} Left: Given an input image, the encoder-decoder network infers a semi-implicit volumetric representation $(S_o, S_f)$ and a camera pose $v$. The semi-implicit volume is then projected from the predicted camera pose to obtain foreground image and mask. The semi-implicit volume is also projected from a novel view $v'$. The projections are required to both match the 2D observation and appear realistic. Right: Projection process mimicking ray marching. 
    % maps it to a latent variable $z$ and infers the camera pose $v$. The latent variable is then decoded to a semi-implicit volumetric rerepsentation, consisting of occupancy $S_o$ and implicit feature $S_f$. 
    % We project both implicit feature and occupancy to obatin foreground images and masks. The rendered images are required to match the observed foreground and masks while 
    }
    \vspace{-3mm}
    \label{fig:vol}
\end{figure*}

\noindent\textbf{Neural renderer for view synthesis and 3D representation.}  Neural rendering \cite{tewari2020state} is a rapidly emerging field aiming to generate photo-realistic images or videos in a controllable manner by rendering its internal 3D representation with various geometric inspirations. 
In particular, many works \cite{lombardi2019neural, mildenhall2020nerf, sitzmann2019deepvoxels,meshry2019neural} present an approach to render a representation based on volumetric feature. 
While the initial applications are in multi-view settings, subsequent works \cite{sitzmann2019scene, nguyen2019hologan} have shown inference of such 3D representation from a single image.
More relevant to our work, HoloGAN \cite{nguyen2019hologan} presents a way to infer this from single single image in an unsupervised setting by generating realistic novel views. However, their approach does not recover an explicit 3D representation. While we also adopt a neural renderer to render our volumetric representation, we explicitly incorporate reasoning about occupancy to improve 3D estimation.

% The internal representation could be built from multi-view \cite{thies2018ignor} or inferred from single-view \cite{kato2019learning}. It could adopt explicit geometry \cite{liu2019soft, kato2018neural, chen2019learning} or implicit feature  \cite{ matusik2003data, nguyen2019hologan, sitzmann2019deepvoxels}. It could be volume-based \cite{nguyen2019hologan, lombardi2019neural}, implicit field \cite{mildenhall2020nerf, sitzmann2019scene}, multi-plane images \cite{meshry2019neural, xu2019deep}  or meshes \cite{zhu2018visual, wang2018pixel2mesh, pan2019deep}. We encourage readers to refer to the recent review \cite{tewari2020state}. 
% which are computational friendly to learning framework but memory inefficient, or meshes \cite{zhu2018visual, wang2018pixel2mesh, pan2019deep} which are expressive and flexible to render but suffer from initialization and registration.
% In our work, we use rendering techniques to circumvents 3D annotations. But rather than aiming for photo-realism or generation controllability, we focus on faithfully uncovering the underlying 3D from image collections.  

\section{Method}

\vspace{-2mm}
Our goal is to learn a model that, given an input image segmented with object mask, outputs a 3D shape in the form of a triangle mesh with texture and the corresponding camera pose. We use a two-step approach. First, we predict a canonical-frame volumetric representation and a camera pose to capture the coarse 3D structure which is consistent with categorical priors. We then convert this coarse volume to a memory-efficient mesh representation which is refined to better match the instance-level details.
 
We propose to learn the category level model from image collections using geometric and adversarial meta-supervision signals. More specifically, our key insight is that the projection of the predicted 3D should explain the observed images and masks, while also appearing realistic from a novel view. 

\vspace{-3mm}
\subsection{Volumetric Reconstruction Model}
\label{sec:method:vol}
\vspace{-2mm}
First we define the volumetric reconstruction model which is learned separately for each category. Given an image  this model predicts a volumetric representation in a canonical frame with corresponding camera pose. Note that unlike approaches that use a deformable category-level shape space, a volumetric representation allows us to capture larger shape and topology variations.

Concretely, we adopt a semi-implicit representation comprised of an explicit occupancy grid $S_o$, with an implicit 3D feature $S_f$, i.e. $S=(S_o, S_f)$. The latter can help capture appearance, texture, material, lighting, etc. This allows synthesizing both mask and appearance from a query view, and thereby lets us use both RGB images and foreground masks as supervision. The overall method is depicted in Figure \ref{fig:vol}.

\noindent \textbf{Encoder-Decoder Architecture.} We learn an encoder-decoder style network $\phi$ to predict this semi-implicit representation $(S, v) \equiv \phi(I)$ where $v$ is the camera pose. The encoder maps the input image to a low-dimensional latent variable $z$ and predict the camera pose, i.e. $(z, v) \equiv \phi_E(I)$. The latent variable  $z$ is then decoded to the volumetric representation, $S\equiv \phi_D(z)$. The key here is that the view-independent decoder learns to predict the shape in a canonical pose across all instances in the category. To further regularize the network, we leverage the observation that many objects exhibit reflection symmetry, and enforce a fixed symmetric plane ($x=0$) via averaging predicted features in symmetrically related locations.
 
\noindent \textbf{Volumetric Rendering.} Our goal is to supervise the volumetric model using only 2D observations.
Therefore, what we need is a rendering function ($\pi$) which projects volumetric representation to obtain 2D images and masks from a query view i.e. $(I, M) \equiv \pi(S_f, S_o, v)$. Similar to other volumetric neural renderers \cite{lombardi2019neural, drcTulsiani19}, we use a geometrically informed projection process by mimicking ray marching. 
%For each pixel, we shoot a ray from camera through pixel center, and composite samples along the ray to obtain the pixel value.

For a given pixel $\mathbf{p}$, we use $D$ samples  along the ray to obtain a `rendered' feature and mask value. Let us denote the coordinate of the $d$-th sample on the ray as $C_v + \lambda_d e_{\mathbf{p}}$ where  $e_{\mathbf{p}}$ is the corresponding ray direction. We sample both implicit feature and occupancy at these locations, denoted as $S_f[C_v + \lambda_d e_{\mathbf{p}}]$ and  $S_o[C_v + \lambda_d e_{\mathbf{p}}]$. We then composite these samples to obtain a per-pixel feature $s^\mathbf{p}_f$ and mask $s^\mathbf{p}_m$, by using the expected value with respect to ray stopping probability \cite{drcTulsiani17}:
\begin{align*}
% continuous
% f_{ij} &= \int_{\lambda_{near}} ^{\lambda_{far}} p[C+\lambda e_{ij}] S_f[C + \lambda e_{ij}] d\lambda, \\
% p[C+\lambda e_{ij}] &= \int_{\mu_{near}}^{\lambda} 1-S_o[C + \mu e_{ij}]d\mu
% discrete 
s^\mathbf{p}_f = &\sum_{d=1} ^{D} (S_o[C + \lambda_d e_{\mathbf{p}}]  \prod_{h=1}^{d-1} (1-S_o[C + \lambda_h e_{\mathbf{p}}]) ) \\& \cdot S_f[C + \lambda_h e_{\mathbf{p}}]
% & p[C+\lambda_k e_{ij}] 
\end{align*}
% where $e_{ij}$ represents the unit vector pointing from camera to the center of pixel $ij$. 
The pixelwise mask value $s^\mathbf{p}_m$ is similarly rendered by setting $S_f$ to constant $1$. While this process lets us directly compute the rendered mask $M$, we use a few upconvolutional layers to transform the rendered 2D feature image to the output color image.

\noindent \textbf{Training.} We supervise this network with only unannotated images and foreground masks. We use three different kinds of supervision (or terms in the loss function):

\noindent \textit{Pixel consistency loss.} Our first term is the simplest one. Any predicted volumetric representation when rendered in the same camera view should explain the input image and mask. This is performed in color space, mask space \cite{liu2019soft}, and perceptual space \cite{zhang2018perceptual}
. 
\begin{align*}
    \mathcal L_{rgb} &= \|\hat I -I\|_1 \\
    \mathcal L_{mask} &= 1 - \frac{\|\hat M\otimes M\|_1}{\|\hat M \oplus M - \hat M \otimes M\|_1} \\ 
    \mathcal L_{perc} &= \|h(\hat I)-h(I)\|_2^2
\end{align*}
%  $$ \mathcal L_{rgb} = \|\hat I -I\|_1; ~~~~~~ \mathcal L_{mask} = 1 - \frac{\|\hat M\otimes M\|_1}{\|\hat M \oplus M - \hat M \otimes M\|_1}; ~~~~~~\mathcal L_{perc} = \|h(\hat I)-h(I)\|_2^2$$
where $\hat I, \hat M$ are rendered image and mask; $h$ is the feature extracted by a pretrained AlexNet \cite{krizhevsky2012imagenet} and $\oplus \otimes$ are element-wise summation and multiplication respectively.

\noindent \textit{View synthesis adversarial loss. } A degenerate solution could arise such that the shape is only plausible from the predicted view. 
To avoid it, we require the projection of predicted shape to appear realistic from a random novel view.
% look like images in the training set. the predicted shape should look plausible from a novel view -- 
Specifically, we sample another camera pose from a fixed prior to render the novel view:  $I' = \pi(\phi_D(z), v'), v'\sim p(v)$. We then present this generated image to an adversarial discriminator with an objective to fool it. We similarly  encourage photo-realism when rendering from the  predicted camera pose. The loss is minimized in a vanilla GAN \cite{goodfellow2014generative} scheme.
$$
L_{adv} = \log \mathcal{D}(I) + \log (1-\mathcal D (\pi(S, v)) + \log (1-\mathcal D (\pi(S, v'))
$$

\noindent \textit{Content consistency loss. } To further regularize the network we build on a insight that the encoder and decoder networks should be self-consistent. Given a synthesized image from the decoder, the encoder should predict the actual content (latent variable with camera pose) that generated that image. 
% To further regularize the network, we enforce that the encoder-decoder to be self-consistent -- the predicted content should be consistent with the actual content to generate the images. 
Formally, 
\begin{align*}
    \mathcal L_{content} = \|\phi_E(\pi_S(S, v))-(z,v) \|_2^2 \\ + \|\phi_E(\pi_S(S, v'))-(z,v')\|_2^2
\end{align*}
Empirically, we found $\mathcal L_{content}$ important to stablize training. 

\textbf{Optimization.} The neural renderer and decoder are trained to minimize all of the above objectives. But the encoder is not optimized with  the adversarial loss, as in \cite{larsen2015autoencoding}.

\begin{figure}
    \centering
    \includegraphics[width=.8\linewidth]{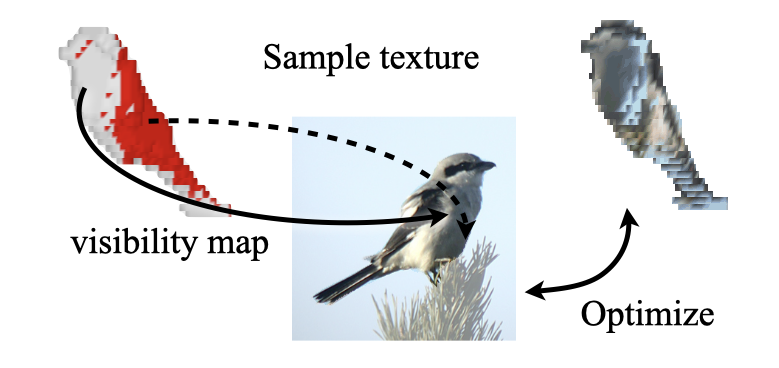}
    \caption{\textbf{Instance-level specialization.}  We convert the inferred volumetric occupancy to an initial mesh. 
    % The mesh vertices are associated with textures sampled from the image.
    The mesh geometry and textures are then iteratively refined to better match the given input.}
    \label{fig:my_label}
    \vspace{-5mm}
\end{figure}
\subsection{Instance-Level Specialization}
\label{sec:method:mesh}
The volumetric representation captures general category-level structure to hallucinate a full 3D shape. However, this shape is coarse as it is: a) limited by the volumetric resolution, and b) generated only from a low-dimensional latent variable. On the other hand, a mesh representation is more flexible and can allow capturing the finer shape details. We therefore go beyond this coarse volumetric prediction, and capture details specific to this instance by converting the volume to an initial mesh, which is then adjusted to better match the input image.

\noindent {\bf Volumetric to Mesh.} We first obatin an initial mesh from the predicted volumetric occupancy. This is done similar to Mesh-RCNN \cite{gkioxari2019mesh} by binarizing the occupancy grid $S_o$ and extracting its surfaces. Next, every vertex is projected to the image to obtain visibility and texture at the vertex. At this step we leverage the symmetry of the mesh to fuse the textures from its reflective symmetric vertex. The final associated texture for each vertex is an average of itself and its visible symmetric neighbors.

%\textbf{Initialization. } 
%Let us denote the mesh as $G = \langle X, T, E\rangle$ where $X$ is the vertex location in space, $T$ the texture  associated with each vertex, $E$ the mesh edges. We initialize the mesh geometry $\langle X^0, E^0 \rangle$ from volumetric representation. Following the  % Every cubic cell  within the surface is converted to a eight vertices, eighteen edges, twelve triangle faces mesh. Then the interior and shared edges and faces are removed, following Mesh-RCNN \cite{gkioxari2019mesh}. %This operation follows operation in Mesh-RCNN \cite{gkioxari2019mesh} and can be processed in batch. 

%\textbf{Mesh texture sampling. } Given the vertex locations, we associate each vertex with a corresponding texture from the input image. Specifically, for every vertex, we first obtain its visibility together with its texture by sampling the pixel of the images at the location where the vertex is projected. Furthermore, we leverage the symmetry of the mesh to fuse the textures from its reflective symmetric vertex. The final associated texture for each vertex is an average of itself and its symmetric neighbors. 

\noindent \textbf{Mesh refinement. }
We optimize the geometry and refine the texture of the mesh iteratively. Given a posed textured input mesh, we first optimize the vertex location and the camera pose such that the projection of the mesh matches the observation. After every step of mesh geometry update, vertex textures are re-sampled from the image given the adjusted projected location. More specifically, we use a mesh-based differential renderer \cite{liu2019soft} to project and render.  The rendered images and masks $(\hat I, \hat M) \equiv \pi_G(G, v)$ are encouraged to be consistent with the input image and foreground mask. We regularize the optimization by penalizing large vertex displacement  $\|\delta X\|_2^2$ and encourage Laplacian smoothness $\|\Delta X\|_2^2$. 

% The  update of mesh geometry and texture sampling are repeated iteratively for every optimization step.
% Every optimization step outputs the displacement of each vertices and camera pose.  
% learn a mesh refinement network $\psi$ takes in a mesh with associated feature $G^k = \langle  X^k, T^k ,E\rangle$ and outputs the displacement of each vertices $\delta X^k$.  In our method, this refinement is performed via a graph neural network \cite{kipf2016semi}.  We repeat this refinement process for certain iterations while resampling the vertex features given the adjusted location $X^{k+1} = X^k + \delta X^k$, and 
% Specifically,
% $
% \Delta x_i = \psi((x_i,g_i, l_i), POOL\{(x_j, g_j, l_j)\}_{j\in \mathcal{N}(i)})
% $

% To supervise the refinement network, we project and render the mesh with feature $G^k$ using a neural renderer $\pi_G$. First, we learn a texture generator that transform the feature $T^k$ to texture for each vertex. Then the textured meshes are rendered to masks and images from the predicted view by a mesh-based differential renderer~\cite{liu2019soft}. 
% $$
% \mathcal L = \sum_{k=1}^K\|\pi_G(G^k, v) - (I, M)\|_1
% $$
% We set $K$ to 5 in our experiments. To further regularize our network, we also add a penalty for large displacement $\|\delta X^k\|_2^2$ and encourage Laplacian smoothness $\|\Delta X^k\|_2^2$.

\section{Experiments}
Our goal is to highlight how our approach learns to predict 3D meshes from image collections in the wild. Specifically, we show 3D reconstruction for 50 object categories from OpenImages dataset~\cite{OpenImages}. Note that this diverse set of reconstructions is an order of magnitude larger than those of any existing approaches. 

However, there is no ground truth for OpenImages. Also, most baseline approaches fail to work on uncurated image collections. In order to provide comparisons, we perform two additional experiments.  First, we compare on data drawn from 3D Warehouse\cite{warehouse}, using rendered images as image collection. Using synthetic data allows us to provide quantitative evaluation and perform ablative analysis. Second, we also compare with some of the other curated common datasets used in the literature (CUB, Chair-in-Wild, ImageNet Quadrapeds). This helps us to qualitatively compare with some baseline approaches.

\begin{figure}
    \centering
    \includegraphics[width=\linewidth]{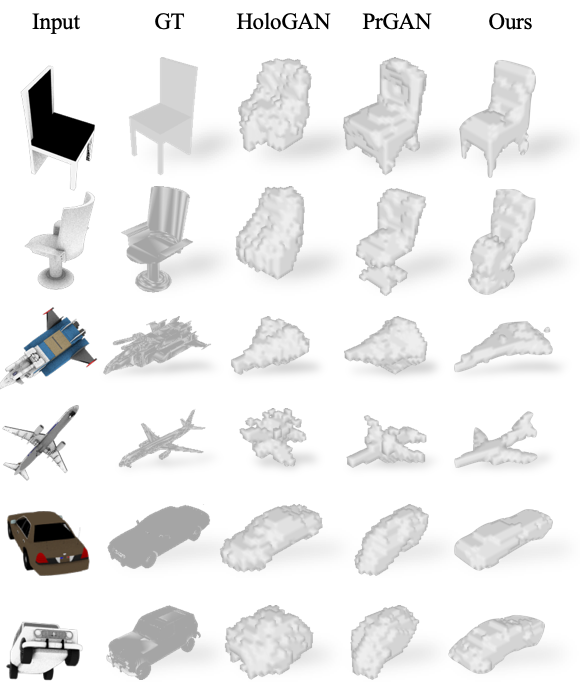}
    \caption{Visualization of categorical volumetric representation across different methods. }
    \label{fig:vis_sn}
\end{figure}

% \begin{table}[]
% \footnotesize
% \centering
%     \begin{tabular}{cccc}
%     \toprule
%      & aero & car & chair \\
%      \midrule
% HoloGAN \cite{nguyen2019hologan}  & 0.28 &	0.43 &	0.26 \\
% PrGAN \cite{gadelha2017prgan} & 0.29 &	0.48 &	0.28 \\
% Ours & \textbf{0.45} &	\textbf{0.59} &	\textbf{0.31} \\
% \bottomrule
%     \end{tabular}
% \caption{Quantitative results on ShapeNet comparing different methods for volumetric reconstruction. }
% \label{tab:quant_vox}
% \end{table}

\begin{table}[]
\footnotesize
\centering
\caption{Quantitative results (3D IoU / F-score) on synthetic data comparing different methods for shape reconstruction. }
    \begin{tabular}{cccc}
    \toprule
     & airplane & car & chair  \\
     \midrule
HoloGAN \cite{nguyen2019hologan}  & 0.28/ 0.31 &	
0.43 /	 \textbf{0.44} & 
0.26 /  0.25 \\
PrGAN \cite{gadelha2017prgan} & 0.29/ 0.18 &
0.48 /  0.37 &	
0.28 /  0.28 \\
\midrule
Ours & \textbf{0.33} /  0.46 &
\textbf{0.55} /  0.43 &	
\textbf{0.31} / 0.29 \\
Ours (refined) & ---- / 	\textbf{0.49} &
---- / 0.42 &	
---- /  \textbf{0.31}\\
\bottomrule
    \end{tabular}
\label{tab:quant_vox}
\vspace{-3mm}
\end{table}

% \section{Experiment 1: ShapeNet-3}
\section{Experiment 1: Synthetic Data}
We first evaluate our method on models from 3D Warehouse~\cite{warehouse}, using the subset recommended by Chang \etal~\cite{chang2015shapenet}. We select three categories which are commonly used to evaluate single-view reconstruction: aeroplane, car, and chair. Note that within a category, the shapes across instances can have a large variation and even different topology, especially for chairs. Each 3D model is rendered from 20 views, with uniformly sampled azimuth $[\ang{0}, \ang{360}]$ and elevation elevation $[\ang{-60}, \ang{60}]$. However, the network is not provided with multi-view associations in training.

\noindent \textbf{Evaluation metrics. } We report 3D IoU with resolution $32^3$ and F-score in the canonical frame for volumetric reconstruction and report F-score \cite{tatarchenko2019single} for mesh refinement. The F-score can be interpreted as the percentage of correctly reconstructed surface.
As our predictions (and those of baselines) can be in an arbitrary canonical frame that is different from the ground truth frame, we explicitly search for azimuth, elevation for each instance and binarizing threshold for each category to align the predicted canonical space with the ground-truth.  

\noindent \textbf{Baselines.} We compare our approach to  \cite{nguyen2019hologan, gadelha2017prgan}. We adapt 
HoloGAN \cite{nguyen2019hologan} by training their system on our data, and obtaining a 3D output by adding a read-off function from the learned volumetric feature to occupancy by enforcing the reprojection consistency with foreground masks. We implement PrGAN \cite{gadelha2017prgan} using our encoder-decoder network. Our implementation provides a boost to original PrGAN. % Finally, we obtain the results on  airplane category for PlatoGAN directly from \cite{platoGAN}.

Figure \ref{fig:vis_sn} visualizes the reconstructions in a canonical frame on 3 categories. HoloGAN is able to reconstruct a blobby shape, but as it does not explicitly represent 3D occupancies, it struggles to generate a coherent shape. PrGAN is able to capture the coarse shape layout but it lacks some details like flat body of aeroplanes. In contrast, we reconstruct the shape more faithfully to the ground-truth as we leverage information from both appearance and foreground masks. Quantitatively, we report the 3D IoU on these categories in Table \ref{tab:quant_vox} and, consistent with the qualitative results, observe empirical gains across all categories.

\begin{figure}
    \centering
    \includegraphics[width=\linewidth]{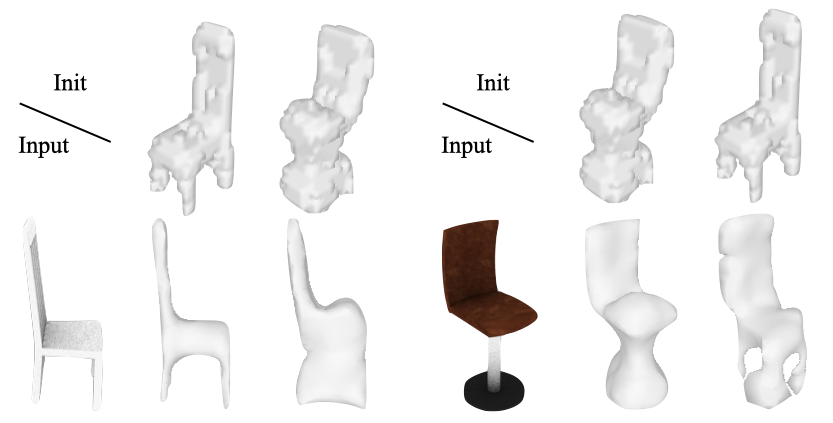}
    \caption{Ablation study: refining mesh initialized from the predicted volume (col 2/5) and another volumes (col 3/6).}
    \vspace{-3mm}
    \label{fig:refine_sn}
\end{figure}
\begin{figure*}
    \centering
    \includegraphics[width=\linewidth]{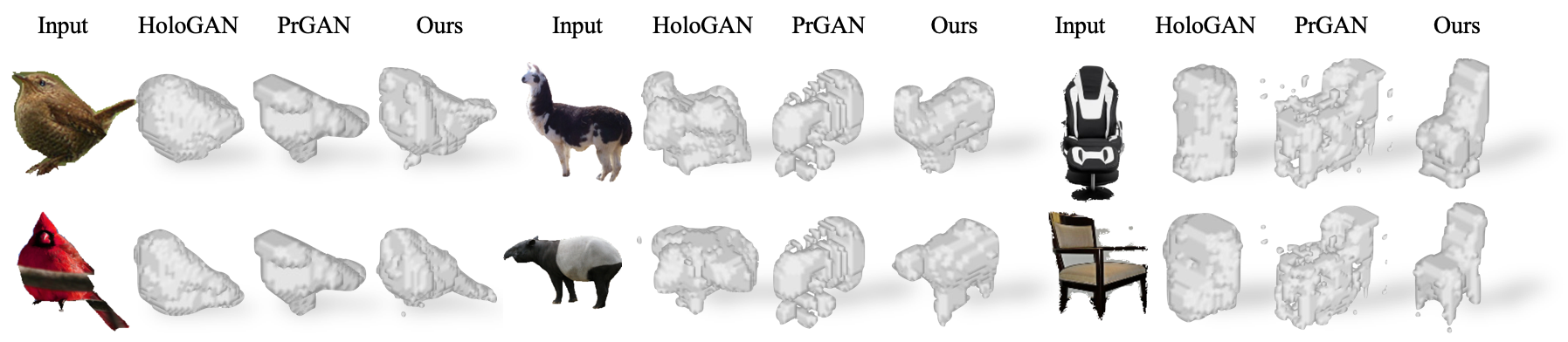}
    \caption{Visualizing categorical volumetric representation across different methods  on CUB-200-2011, Quadrupeds, Chairs in the wild. }
    \label{fig:vis_realbaseline}
    \vspace{-3mm}
\end{figure*}
\begin{table}[]
\footnotesize
\centering
\caption{Quantitative results (3D IoU) on synthetic data to ablate the effect of each loss term. }
\vspace{1mm}
    \begin{tabular}{cccc}
    \toprule
     & airplane & car & chair  \\
     \midrule
Ours & \textbf{0.33} &
\textbf{0.55} &	
\textbf{0.31} \\
Ours  $-\mathcal L_{adv}$ & 0.25 & 0.44 & 0.22 \\
Ours $-\mathcal L_{cont}$ & 0.24 & 0.54 & 0.23 \\

\bottomrule
    \end{tabular}
\label{tab:ablate_loss_quant}
\vspace{-5mm}
\end{table}

% \begin{figure}[b]
% \vspace{-3mm}
% \begin{floatrow}
% % \MyBox{
% \ffigbox[10cm]{%
%   \includegraphics[width=.9\linewidth]{exp/ablation.png}
% }{%
%   \caption{Ablation: the 1st row visualizes the reconstructed shape from the predicted view. The 2nd row shows the shape from a novel view.}%
%     \label{fig:ablation}
% }

% \capbtabbox{%
%     \begin{tabular}{cccc}
%     \toprule
%      & aero & car & chair \\
%      \midrule
% Ours & 0.45 & 0.59 &	0.31 \\
% Ours  $-\mathcal L_{adv}$ & 0.25 & 0.44 & 0.22 \\
% Ours $-\mathcal L_{cont}$ & 0.24 & 0.54 & 0.23 \\
% % Ours $-\mathcal L_{cont} + \mathcal L_{pose}$  & 0.35 & 0.57 & 0.31 \\
% \bottomrule
%     \end{tabular}
% }{    
%     \caption{Quantitative results on ShapeNet to ablate the effect of each loss term. }
%     \label{tab:ablation}
% }
% \end{floatrow}
% \end{figure}

\noindent \textbf{Mesh Refinement.}
Table \ref{tab:quant_vox} also reports the evaluation of the mesh refinement stage.  Compared with the initial meshes converted from volumetric representation, our specialized meshes match the true shape better.

In  Figure \ref{fig:refine_sn}, we visualize the refinement results with an interesting ablation to further highlight the importance of mesh initialization. Instead of initializing with our predicted volume, we initialize the mesh from another chair consisting of different numbers of chair legs. The refinement fails to specialize well. This indicates that the meshes for all instances cannot be adjusted from one single shape especially when shapes have a large variance, and that our volumetric prediction, though coarse, provides an important initialization for the instance-level refinement.

\begin{figure}
    \centering
    \includegraphics[width=.8\linewidth]{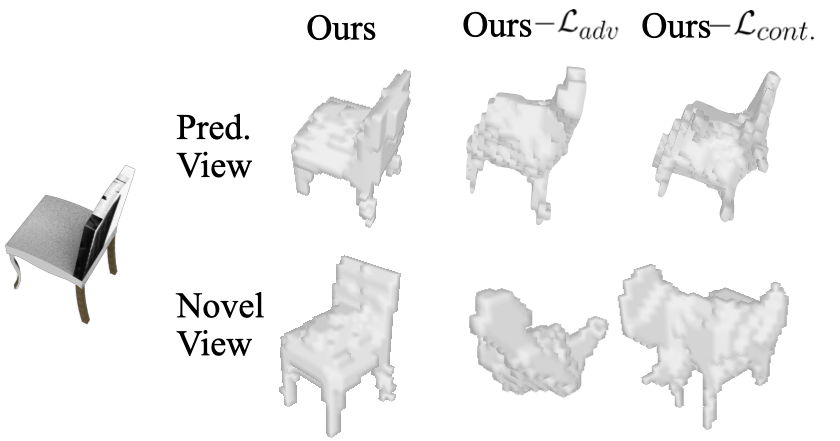}
    \caption{Ablation study: comparing reconstructed volumes when the model disables different loss terms.}
    \label{fig:ablation_loss}
    \vspace{-3mm}
\end{figure}

\noindent \textbf{Loss ablation. } We provide quantitative (Table \ref{tab:ablate_loss_quant}) and qualitative (Figure \ref{fig:ablation_loss}) results to show each loss term is necessary. Without adversarial loss, the model collapses to generate shapes only looking similar to the input from the predicted view. It does not even look like a chair from another view, since this degenerate solution is not penalized by other losses.  Without content loss, the performance also drops, especially on categories with larger shape variance like chairs. The consistency loss is not ablated because it is needed for the task of reconstruction.

%\begin{itemize}
%    \item HoloGAN \cite{nguyen2019hologan}. This approach learns a 3D feature representation but does not explicitly reason about occupancy.  Additionally, their proposed rendering operation does not explicitly enforce occlusion when projecting the implicit feature to 2D. We train their system on our data, and obtain a 3D output by adding a read-off function from the learned volumetric feature to occupancy by enforcing the reprojection consistency with foreground masks.
%    \item PrGAN \cite{gadelha2017prgan}. While our approach uses both shape and appearance synthesis for supervision, the approach from ~\cite{gadelha2017prgan} only uses foreground masks. To compare this supervision objective with ours, we adapt our objective to only include foreground mask related terms. 
%\end{itemize}

\begin{figure*}
    \centering
    \includegraphics[width=\linewidth]{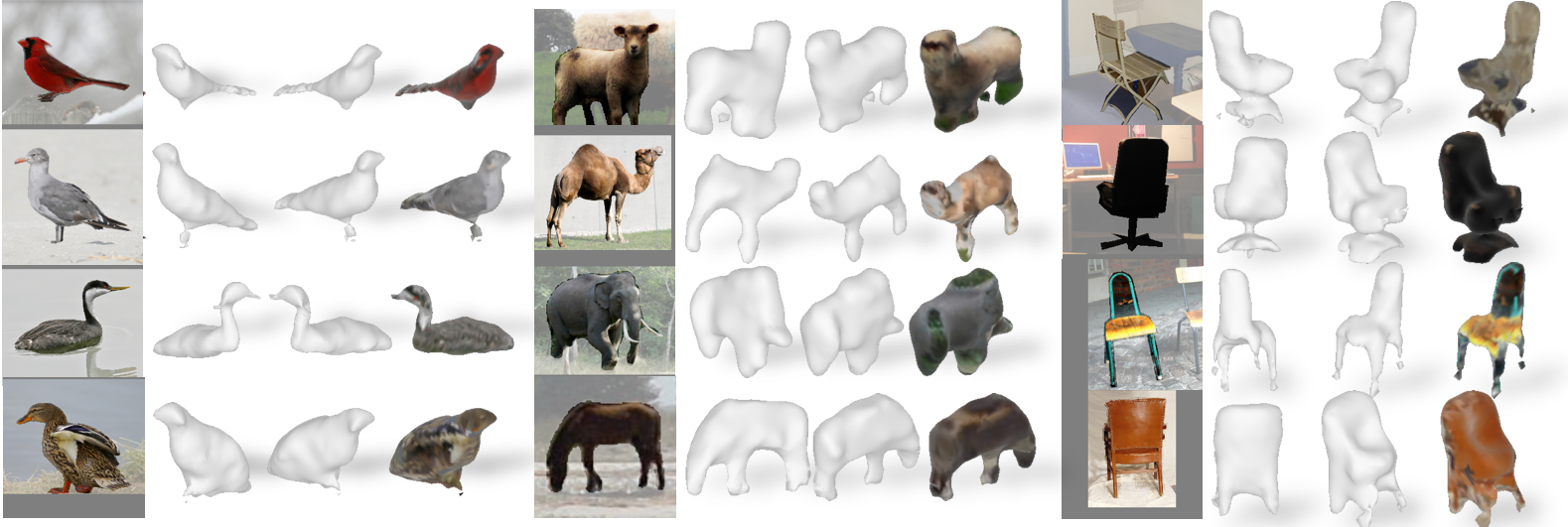}
\caption{Visualizing our refined shapes from the predicted view (2nd column in each quadruplet) and a novel view (3rd+4th / quadruplet) on CUB, Quadrupeds, Chairs in the wild. We are able to capture both the shared shapes in categories and instance-specific differences.  }
    \label{fig:vis_cub}
\end{figure*}

\section{Experiment 2: Curated Collections}
We also examine our method on three real-world datasets that have been curated and used in the literature for the 3D reconstruction problem: 

\noindent {\bf CUB-200-2011} \cite{wah2011caltech}: The CUB dataset consists of 6k images of 200 bird species with annotated foreground masks. 

\noindent {\bf Quadrupeds from ImageNet} \cite{deng2009imagenet}: The Quadrupeds dataset consists of 25k images of different quadrupeds from ImageNet. Masks are provided by Kulkarni \etal~\cite{kulkarni2020acsm}, who use an off-the-shelf segmentation system \cite{kirillov2019pointrend} and manually filter out the truncated or noisy instances. Quadrupeds consists of multiple 4-legged animal species including buffalo, camels, sheep, dogs, etc. The animals also exhibit rich articulation \eg running, lying, heads up or heads down. This makes the underlying shape variance significantly larger than the CUB dataset.

\noindent {\bf Chairs in the wild \cite{deng2009imagenet, xiang2014beyond, oh2016deep}}: For chairs in the wild, we combine chairs in PASCAL3D, ImageNet, and Stanford Online Products Dataset  to get 2084 images for training and 271 for testing. Foreground masks in \cite{deng2009imagenet, oh2016deep} are from segmentation systems \cite{kirillov2019pointrend, chen2017deeplab} and those in \cite{xiang2014beyond} are from annotations.

Figure \ref{fig:vis_realbaseline} qualitatively compares our volumetric reconstruction to baselines on the 3 separate real-world datasets.  Similar to results on synthetic data, HoloGAN reconstructs only coarse blobby volumes as it does not explicitly consider  occupancy or geometric-informed projection. PrGAN collapses to shapes with little variance, since it does not use  appearance cues. But the real datasets have noisier foreground masks and textures contain more information. In contrast, we are able to learn the coarse categorical shape just from the foreground images. Our reconstructions also capture subtle differences like the length of bird tails, articulated heads of the quadrupeds, the style of chairs. 

\noindent \textbf{Mesh Refinement. } Figure \ref{fig:vis_cub} visualizes  our refined meshes from the predicted view and a novel view on these three dataset. We observe that we predict meaningful texture even for invisible regions  and that the shape of the mesh also looks plausible from another view.
On CUB-200-2011, our method captures the categorical shapes like blobby bodies, beaks and tails while captures subtle shape differences between birds such as the tail length, body width, neck bending, etc.  On Quadrupeds, we are able to capture quadrupeds common traits such as torso with one head and front back legs. We can  also depict their uniqueness such as the camel hump and longer legs, the tapir having stout neck, the sheep raising up its heads, the horse bending down its neck, \etc. On Chairs in the wild,  the learned common model differentiates one-leg and four-leg chairs respectively. The four legs and seat can be hallucinated even when occluded. The subtle differences such as a wide or a narrow chair back are also captured.
Despite the challenges in the datasets, it is encouraging that our model can capture both, the common shapes and specialized details just by learning from these unannotated image collections.

\section{Experiment 3: OpenImages 50 Categories}
\begin{figure*}[]
    \centering
    \vspace{-3mm}
    \includegraphics[width=\linewidth]{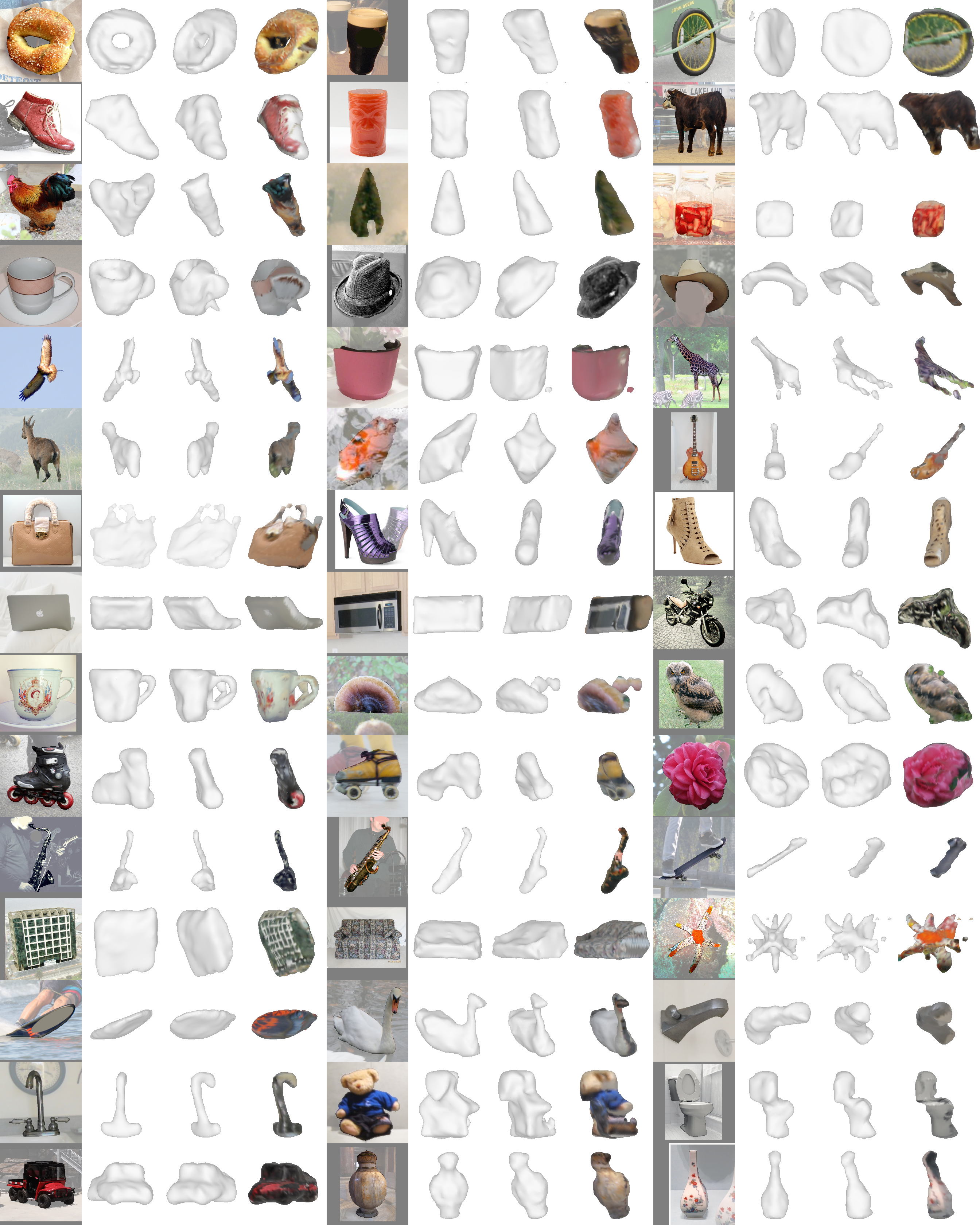}
    \caption{Visualizing our reconstructed meshes from the predicted view (2nd in each quadruplet) and a novel view (3rd and 4th in each quadruplet) trained on multiple categories on Open Images. }
    \label{fig:openimages}
\end{figure*}

Finally, the highlight of our model is the ability to scale to images in the wild. We evaluate our model on 50 categories on Open Images including bagel, water tap, hat, \etc. The size of each category ranges from 500 to 20k. The foreground masks \cite{OpenImagesSegmentation} are from annotation and filtered by a fine-tuned occlusion classifier. Figure \ref{fig:openimages} visualizes the reconstructed meshes from the predicted view and a novel view. Our method works on a large number of categories, including thin (water taps, saxophone), flat (wheels, surfboards),  blobby structures (Christmas trees, vases). We are able to reconstruct shapes with various topology such as bagels, mugs, handbags. The model captures the categorical shapes shared within classes and hallucinates plausible occluded regions (mushroom, mugs). We can also captures details at instance-level, such as the number of wheels of roller-skaters, styles of high-heels, hats, \etc. 

\begin{figure}
    \centering
    \includegraphics[width=\linewidth]{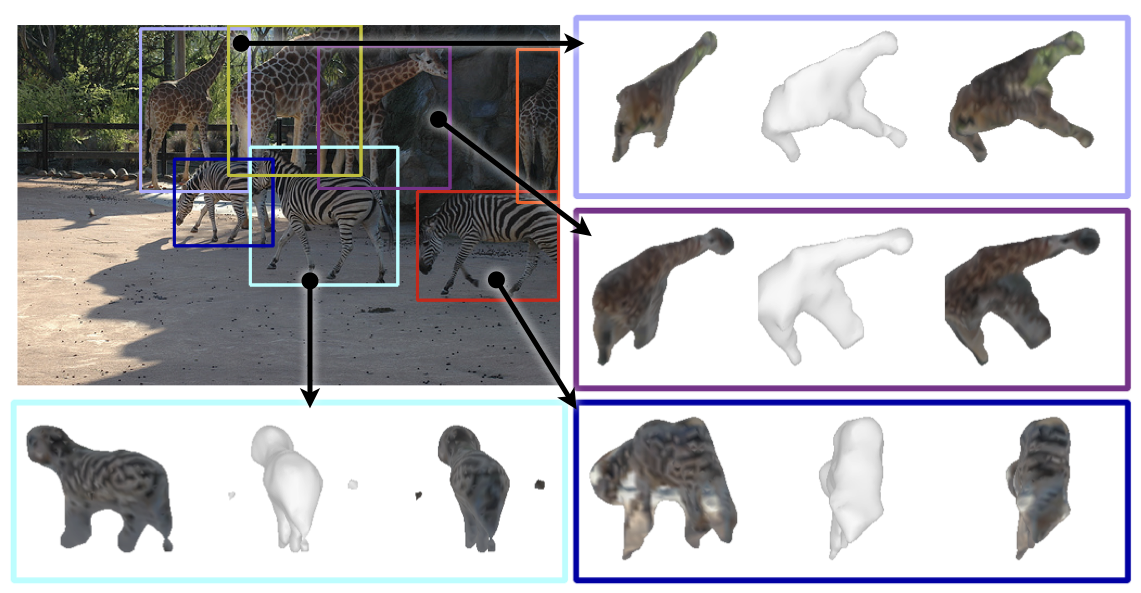}
    \vspace{2mm}
    \includegraphics[width=\linewidth]{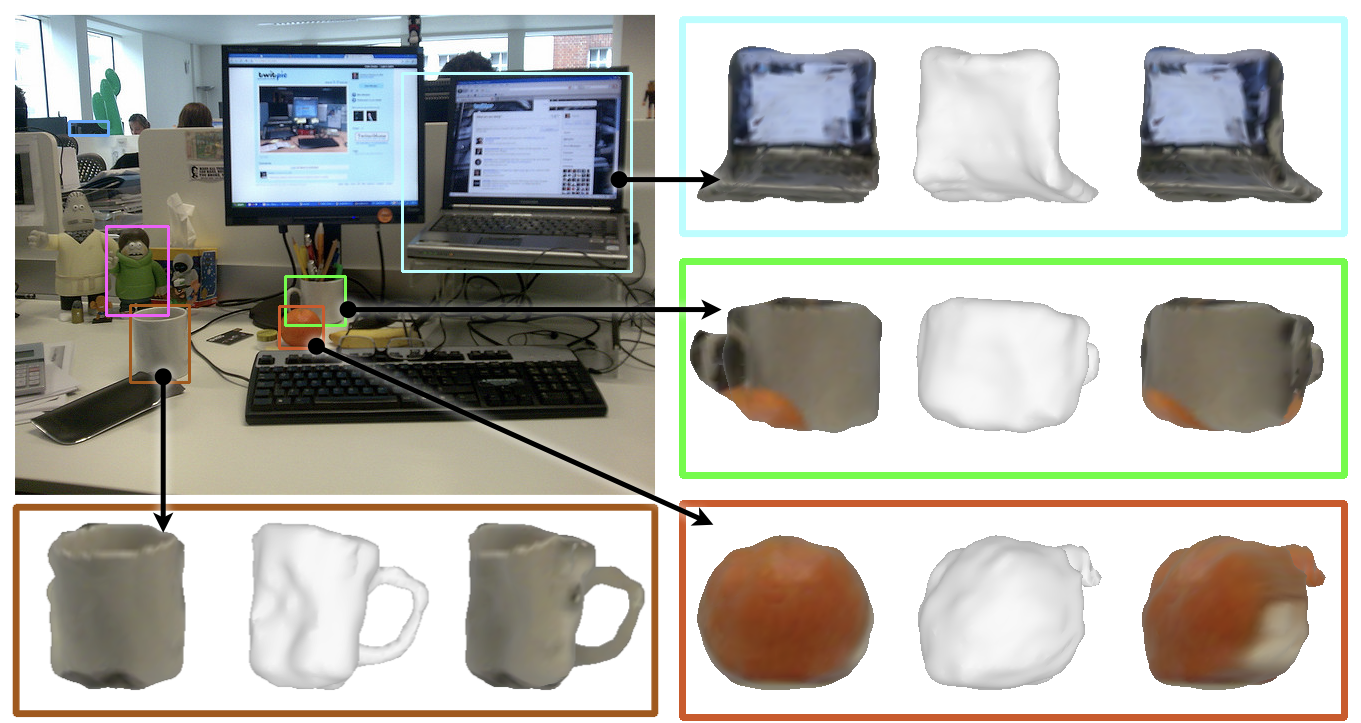}
    \caption{Test on COCO: visualization of lifting detection results to meshes via the shelf-supervised models. }
    \label{fig:coco}
    \vspace{-3mm}
\end{figure}

\noindent \textbf{Integrating on COCO. } We additionally show results of our models on COCO \cite{lin2014microsoft} without fine-tuning (Figure \ref{fig:coco}). We first detect and segment the objects with off-the-shelf segmentation \cite{kirillov2019pointrend} system. Based on the predicted classes, we then pass the segmented objects to our category-specific models which are trained on previous datasets. 
% Some classes in COCO may map to their sub-categories or super-categories due to mismatch (\eg cups in COCO are mapped to mugs in Open Images). 
Despite more cluttered scenes and the dataset domain shift, our models can lift the 2D detection to 3D meshes for various categories while preserving instance details. It is an encouraging step towards scalable 3D recognition in the wild. 


%\input{exp/alation_view}
%\textbf{Assumption of viewpoint distribution.} \todo{supplementary?} We briefly analyze the effect of viewpoint prior. In figure \ref{fig:mug} we visualize volumetric reconstruction training with different viewpoint prior on the mug category of ShapeNet. While our method is robust to some view distribution mismatch, the shapes display artifact (\eg two handles) when the assumed prior is far from the ground-truth viewpoint distribution. It is because different viewpoint distribution may induce different 3D shapes as the adversarial loss matches its projections with the existing image collections. We notice similar artifacts when training on the real datasets (\eg starfish and mugs on OpenImages ), as camera pose biases exist by human photographers (\eg front view of starfish or  mugs with handles). While we assume azimuth from uniform distribution across all experiments and have achieved some promising results on various categories, we encourage more works to explore the direction of better viewpoint distribution prior. 

\section{Conclusion}
We presented an approach to predict 3D representations from unannotated images by learning a category-level volumetric prediction followed by instance-level mesh specialization.  We found that both are important to infer an accurate 3D reconstruction. While we obtained encouraging results across diverse categories, our approach has several limitations. For example, our rendering model is simplistic and not incorporate lighting during rendering. Thus we cannot easily reason about concave structure. Additionally, while we only examined setups without annotated supervision like mesh templates, our system could potentially incorporate additional (sparse) supervision to improve the reconstruction quality.
% It would also be interesting to incorporate correspondence between image and mesh in our system. 
While these additional challenges still remain, we believe our work on inferring accurate reconstruction with limited supervising can provide a scalable basis to achieve the goal of reconstructing generic objects in the wild. 

% cooperate lighting model / convex-concave confusion, improve volumetric prediction leveraging supervision from shading in mesh

% \vspace{2mm}
\textbf{Acknowledgements.}
The authors would like to thank  Nilesh Kulkarni for providing segmentation masks of Quadrupeds. We would also like to thank Chen-Hsuan Lin, Chaoyang Wang, Nathaniel Chodosh and Jason Zhang for fruitful discussion and detailed feedback on manuscript. Carnegie Mellon Effort has been supported by DARPA MCS, DARPA SAIL-ON, ONR MURI and ONR YIP. 

% \clearpage
{\small
\bibliographystyle{ieee_fullname}
\bibliography{references}
}

\clearpage
\begin{strip}\centering
\vspace{-1.4cm}
    \includegraphics[width=\linewidth]{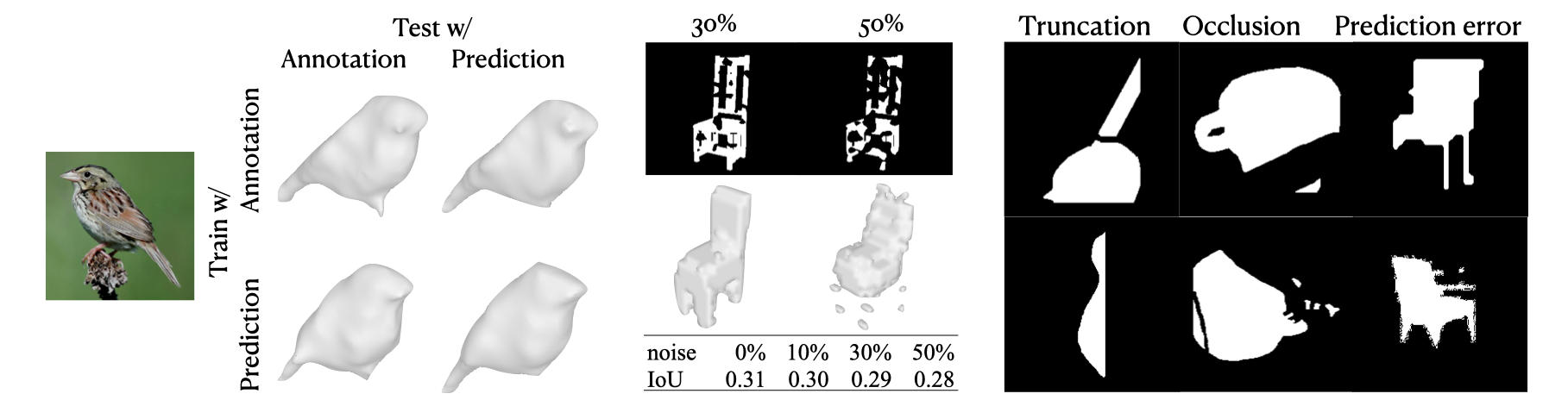}
    \captionof{figure}{Left: qualitative results on CUB replacing annotation with prediction in training and/or test time. Middle: quantitative results on ShapeNet-chairs by adding random noise on masks during both training and inference. Right: masks used to train the models in the paper. }
    \label{fig:det}
\end{strip}

% \begin{figure*}
%     \centering
%     \includegraphics[width=\linewidth]{exp/ablation_det_table.png}
%     \captionof{figure}{Left: qualitative results on CUB replacing annotation with prediction in training and/or test time. Middle: quantitative results on ShapeNet-chairs by adding random noise on masks during both training and inference. Right: masks used to train the models in the paper. }
%     \label{fig:det}
% \end{figure*}

\section{Ablation Study}
\begin{figure}
    \centering
    \includegraphics[width=\linewidth]{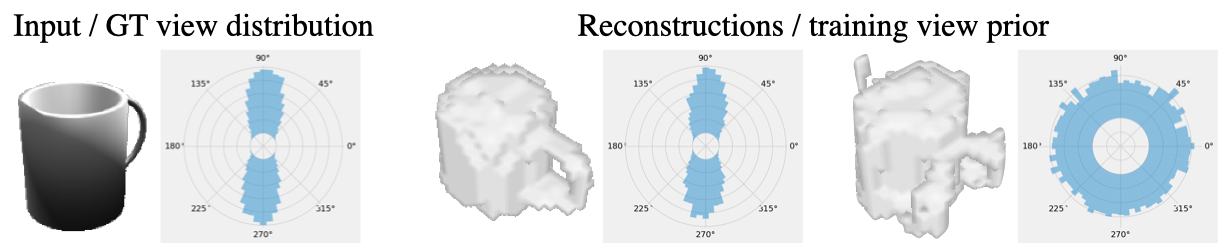}
    \caption{Results on training models with different viewpoint priors.}
    \label{fig:mug}
\end{figure}
\noindent\textbf{Assumption of viewpoint distribution.}  We briefly analyze the effect of viewpoint prior. In figure \ref{fig:mug} we visualize volumetric reconstruction training with different viewpoint prior on the mug category of ShapeNet. While our method is robust to some view distribution mismatch, the shapes display artifact (\eg two handles) when the assumed prior is far from the ground-truth viewpoint distribution. It is because different viewpoint distribution may induce different 3D shapes as the adversarial loss matches its projections with the existing image collections. We notice similar artifacts when training on the real datasets (\eg starfish and mugs on OpenImages ), as camera pose biases exist by human photographers (\eg front view of starfish or  mugs with handles). While we assume azimuth from uniform distribution across all experiments and have achieved some promising results on various categories, we encourage more works to explore the direction of better viewpoint distribution prior.

\noindent\textbf{Robustness against segmentation quality.} 
Our model depends on the segmentation quality, as it is the only supervision. We ablate our model with noisy masks, both qualitatively and quantitatively. The model trained/tested with predictions from [19] (left) or with synthesized noise (mid) performs comparably to using GT, until considerably severe corruption.
Our experiments in paper have already suggested that our model is robust to the noise as masks might be  truncated, occluded, or corrupted due to prediction error (Fig \ref{fig:det} right). We also visualized the masks used in the main paper (Fig \ref{fig:det} right). Our experiments suggest that our model is robust to the noise as masks might be  truncated, occluded, or corrupted due to prediction error. 

\section{Architecture Details }
\textbf{Neural Network Architecture. }
The encoder is comprised of 4 convolution blocks followed by two heads to output $v$ and $z$. Each block consists of $Conv(3\times3) \to LeakyReLU$. The feature from the last block is fed to 2 fully-connected layers to get $v$ and is fed to Average Pooling with another fully-connected layer to output $z$. $v$ is in 2-dim to represent azimuth and elevation while the dimensionality of latent variable $z$ is 128. 

The decoder follows StyleGAN\cite{karras2019style}  to use the latent variable $z$ as a ``style" parameters to stylize a constant $256\times 4^3$ feature. Given $z$, the constant is upsampleed to the implicit 3D feature $S_f$ by a sequence of style blocks. Then $S_f$ is transformed to get the occupancy grid $S_o$ by a $3\times 3\times 3$ Deconv layer with Sigmoid activation. Among all of our experiments, our decoder consists of 2 style blocks each of which are built with $Deconv \to AdaIN \to LeakyReLU$. The shape of $S_f$ is $64 \times 16^3$ and the shape of $S_o$ is $1\times 32^3$.

% \textbf{Visibility in Local Feature Sampling. }
% To sample every visible vertex in the current view, we need to calculate visibility of every vertex. 
% We find the closest vertex to each pixel from the predicted view.  This process could be performed fast by a rasterizer. The local feature of invisible vertices are set to zeros, i.e. $ w_i I[\pi(X_i, v)]$ where $w_i$ is a binary variable indicating the visibility of vertex $i$.

\textbf{Training Details. } We optimize the losses with Adam \cite{kingma2014adam} optimizer in learning rate $10^{-4}$. The learning rate is scheduled to decay linearly after $10$k iterations, following prior work \cite{zhu2017unpaired}. We weight the losses such that they are around the same scale at the start of training. Specifically, we use $\lambda=10$ for $\mathcal{L}_{pixel} + \mathcal{L}_{perc}$, $1$ for $\mathcal{L}_{adv}$ and $\mathcal{L}_{content}$.  The volumetric reconstruction network is optimized for $80$k. Due to the diverse appearance and data noise on Quadrupeds, we additionally regularize the network by an L2 distance between the predicted voxels and the mean shape of all quadrupeds. The model can still capture the articulation for different instance.

\section{F-score Calculation}
In order to calculate F-score -- the harmonic mean of recall and precision, the meshes are first converted to point cloud by uniformly sampling from surfaces. The recall is considered as the percentage of ground-truth points whose nearest neighbour in predicted point cloud is within a threshold while the precision is calculated as the other  prediction-to-target way.

% \section{More Results}
% Please refer to the website in the supplementary material. 
\end{document}